# The Effects of Quantum Randomness on a System Exhibiting Computational Creativity

Azlan Iqbal
*College of Computing and Informatics*
*Universiti Tenaga Nasional*
Putrajaya, Malaysia
azlan@uniten.edu.my

*Abstract*—We present experimental results on the effects of using quantum or 'truly' random numbers, as opposed to pseudorandom numbers, in a system that exhibits computational creativity (given its ability to compose original chess problems). The results indicate that using quantum random numbers too often or too seldom in the composing process does not have any positive effect on the output generated. Interestingly, there is a 'sweet spot' of using quantum random numbers 15% of the time that results in fewer statistical outliers. Overall, it would appear that there may indeed be a slight advantage to using quantum random numbers in such a system and this may also be true in other systems that exhibit computational creativity. The benefits of doing so should, however, be weighed against the overhead of obtaining quantum random numbers in contrast to a pseudorandom number generator that is likely more convenient to incorporate.

*Keywords—Quantum, randomness, pseudorandom, creativity, chess*

## I. Introduction

Quantum randomness can be understood as the phenomenon of random numbers generated based on the principles of quantum mechanics [1]. As opposed to pseudorandom numbers generated algorithmically using a computer program, quantum random numbers are said to be 'truly' random or unpredictable. They can be generated using, for example, photons from a laser that become entangled and then measured to produce numbers [2]. While particularly useful in cryptographic protocols [3], the effects of quantum randomness, beneficial or otherwise, in certain other areas of computer science remains largely unexplored.

Computational creativity is a relatively new sub-field of artificial intelligence (AI) that covers with more specificity the ability of computers to generate novel objects of creative value or that humans would find surprising [4-6]. This may include systems that generate music or art [7-8]. While not on the same level as the greatest or even professional musicians or painters, the works produced are often better than what *most* humans might produce and they take less time as well. Furthermore, the artificial works may be subject to some degree of negative bias by humans [9-10].

We could not find any notable and relevant scientific literature that looked into the effects of quantum randomness on computational creativity (or even creativity in humans), so regrettably a review or even an overview of the research area is not possible here. However, with this article, we hope to open the area for further and more detailed exploration by other researchers in the future. In any case, the concept of computational creativity should not be conflated with 'machine learning' because the latter does not necessarily involve the output of objects of creative value to humans.

In machine learning, the use of quantum randomness has been shown to have an inconsistent but significant effect on the output [11]. Once again, research involving quantum randomness and artificial intelligence is sparse at this point and it will likely be several years or decades before anything conclusive (e.g. based on a large number of replicated studies) can be determined. The question we wanted to address in our research is whether the application of quantum randomness made any difference to a computationally creative system which otherwise used pseudorandom numbers (where required in its processing).

## II. The Computationally Creative System

The system we used for testing purposes is the computer program, *Chesthetica*, an automatic composer of chess problems [12]. It incorporates the 'Digital Synaptic Neural Substrate' (DSNS) computational creativity approach that does not use any kind of machine learning [13]. Rather, it combines attributes from different domains (e.g. paintings, music, chess sequences) using stochastic methods in order to derive new feature specifications which can be used to generate new objects in any of the source domains. Interested readers are encouraged to peruse the aforementioned reference for further details which would be too lengthy to repeat here.

*Chesthetica* is able to generate original forced mates in two, three, four and five moves, and also 'studies' which is a longer type of chess problem that does not end in mate but instead a decisively won or drawn position. It does so entirely autonomously, i.e. from beginning to end, with no human intervention in the composing process. The general composing algorithm is provided in 'Appendix A' of [13]. It is quite long and would interfere with the flow of this article so we will not repeat it here. *Chesthetica* uses pseudorandom numbers at certain points in its composing process.

Notably, they are used to select from among the different piece configurations 'permissible' (at a given time) to compose with, to determine the number of composing attempts for each selected configuration, and to determine which pieces to place on available squares or to remove from the board given any constraints that may be in effect. Randomness is therefore a significant yet minor component in the overall composing process which relies more on the attributes derived from the materials from various domains provided to it.

It has been demonstrated, for instance, that composing using the DSNS approach (versus a purely pseudorandom approach) performs better, i.e. produces compositions of higher aesthetic quality, on average [13]. The pseudorandom numbers used by the program are produced by the internal 'random' function of its programming language. For example, the statement, *a = Int(100 \* Rnd)*, will generate a random integer between 0 and 99 and attribute it to variable, *a*. Different computationally creative systems may need to employ randomness for different purposes in different ways using different programming languages either more or less frequently so there will be inherent limits to how precisely experiments in this area can be replicated.

## III. THE EXPERIMENTAL SETUP

To test whether using quantum or 'truly' random numbers instead of pseudorandom numbers in the composing process made any kind of difference, we compared the performance of two different settings of *Chesthetica*. One used the standard pseudorandom number generator whereas the other used quantum random numbers obtained live from the 'ANU Quantum Random Numbers Server'. To quote directly from the website:

"*The random numbers are generated in real-time in our lab by measuring the quantum fluctuations of the vacuum. The vacuum is described very differently in the quantum mechanical context than in the classical context. Traditionally, a vacuum is considered as a space that is empty of matter or photons. Quantum mechanically, however, that same space resembles a sea of virtual particles appearing and disappearing all the time. This result is due to the fact that the vacuum still possesses a zero-point energy. Consequently, the electromagnetic field of the vacuum exhibits random fluctuations in phase and amplitude at all frequencies. By carefully measuring these fluctuations, we are able to generate ultra-high bandwidth random numbers.*" [14]

The *Chesthetica* program was modified to be able to request and obtain these quantum random numbers whenever a random number was required during the composing process. Alternatively, it could rely on its default internal algorithmic pseudorandom number generator. To minimize the number of variables in the experiment, we decided to compare mainly forced mate in three compositions. These compositions would be compared in terms of their mean aesthetic score and also in terms of the quantity generated in a given time period.

The aesthetic score of a chess composition would be calculated using *Chesthetica* itself based on its internal computational aesthetics model which has been demonstrated to correlate positively and well with the assessment of human chess experts [15]. The aesthetics model is too lengthy for this article and it would be redundant to explain again here but suffice to say it uses mathematical formalizations of seven known aesthetic principles (e.g. checkmate economically, violate heuristics successfully) and ten chess themes (e.g. pin, skewer, zugzwang). More details can be found in the aforementioned reference.

The comparisons would involve four sets. The first using only pseudorandom numbers. The second using quantum randomness 5% of the time. This means that a quantum random number would have a 5% chance of being chosen over a pseudorandom number whenever one was required in the composing process. We wanted to test if a small introduction of quantum randomness had any effect at all. The next two sets would involve quantum randomness at 15% and 25%. These are more significant infusions of quantum randomness into the composing process.

The testing would be done in two rounds of four days (96 hours) of composing, or eight days in total, using two computers running 10 instances each of *Chesthetica*. This means each set as just described would have exactly the same hardware and software, i.e. *Intel® Pentium® Silver J5005 CPU @ 1.50 GHz, 4 GB of RAM, Windows 10 Home, 64-bit*. We decided not to test up to 100% or total quantum randomness here because preliminary investigations suggested it was not beneficial to the process or even detrimental in some cases. This was true even at 50%.

## IV. EXPERIMENTAL RESULTS & DISCUSSION

The experimental results based on the setup explained in the previous section are as follows. Table I shows the mean aesthetic scores for each set over four days (96 hours) of running *Chesthetica*.

TABLE I. MEAN AESTHETIC SCORES OF COMPOSITIONS

| Pseudo | Q 5% | Q 15% | Q 25% |
|---|---|---|---|
| 2.330 | 2.313 | 2.315 | 2.274 |

The pseudorandom set scored the highest, on average, followed by the quantum random number sets at 15%, 5% and finally 25%. However, a single factor analysis of variance (ANOVA) test on the data showed no difference of statistical significance between the means of the sets; $F(3, 4090) = 1.814$, $p = 0.142$. The only difference of statistical significance detected was between the pseudorandom set and the 25% set using a two-sample t-test assuming unequal variances (TTUV); $t(1724) = 2.261$, $p = 0.024$. Table II shows the quantities of compositions generated over the same period in each set by each of the 10 instances (#) of the program.

TABLE II. QUANTITIES OF COMPOSITIONS GENERATED

| # | Pseudo | Q 5% | Q 15% | Q 25% |
|---|---|---|---|---|
| 1 | 138 | 151 | 174 | 128 |
| 2 | 5 | 133 | 108 | 114 |
| 3 | 135 | 127 | 120 | 0 |
| 4 | 138 | 73 | 132 | 98 |
| 5 | 126 | 116 | 122 | 131 |
| 6 | 117 | 67 | 122 | 0 |
| 7 | 143 | 123 | 88 | 141 |
| 8 | 117 | 158 | 124 | 0 |
| 9 | 139 | 140 | 125 | 186 |
| 10 | 142 | 121 | 105 | 102 |

The quantities shown likely include a much smaller proportion of forced mates in two and some invalid sequences as well. The former tend to arise as a byproduct of the mate in three composition process and the latter could be due to unpredictable reasons or unknown bugs in the software. We did not want to remove them from the total quantity of compositions actually generated (leaving only mates in three) because this would introduce artificiality into what had otherwise resulted 'organically'. Mates in two are also perfectly usable as outputs of the program whereas the handful of erroneous sequences are probably inevitable given enough processing time.

In any case, the quantities are independent of the aesthetic scores. There is no known way to predict how many problems a particular instance of *Chesthetica* will compose in any given time period given particular hardware and program settings which explains why some instances had zero compositions and others as high as 186. The average quantity of compositions for the pseudorandom, quantum 5%, quantum 15% and quantum 25% sets are 120, 120.9, 122 and 90, respectively. In terms of quantity, an ANOVA test showed no difference of statistical significance between these sets either; $F (3, 36) = 1.271$, $p = 0.299$. Comparing even the 15% set (122) and 25% set (90) using a TTUV, there was no statistically significant difference; quantity is therefore unaffected by the use of quantum random numbers.

So far it would seem that only the average aesthetics is influenced (negatively) by the use of quantum random numbers (over pseudorandom) when the quantum random numbers are employed 25% of the time, but not at the lower frequencies tested. We therefore decided to test for the number of statistical (aesthetic) outliers in each set. This would provide some measure of the consistency of the aesthetics of the compositions generated. Fewer outliers in a set are generally seen as a good thing overall. Table III shows the first quartile (Q1), third quartile (Q3), the interquartile range (IQR), i.e. the difference between Q1 and Q3, the upper bound (UB) and lower bound (LB) of each set.

TABLE III. Aesthetic Score Outlier Determination

|  | Q1 | Q3 | IQR | UB | LB |
|---|---|---|---|---|---|
| **Pseudo** | 1.948 | 2.688 | 0.740 | 3.798 | 0.838 |
| **Q 5%** | 1.935 | 2.655 | 0.721 | 3.736 | 0.854 |
| **Q 15%** | 1.925 | 2.662 | 0.737 | 3.767 | 0.820 |
| **Q 25%** | 1.852 | 2.639 | 0.787 | 3.819 | 0.673 |

Only upper bound outliers were discovered in the sets. Table IV shows these values sorted and italicized.

TABLE IV. Aesthetic Outliers Detected in Each Set

| **Pseudo** | **Q 5%** | **Q 15%** | **Q 25%** |
|---|---|---|---|
| *4.219* | *4.154* | *4.359* | *4.3* |
| *4.055* | *3.984* | *4.023* | *3.959* |
| *4.049* | *3.82* | *3.773* | *3.827* |
| *3.909* | *3.765* | *3.671* | *3.783* |
| *3.835* | *3.698* | *3.671* | *3.782* |
| *3.817* | *3.654* | *3.645* | *3.776* |
| *3.776* | *3.636* | *3.642* | *3.76* |

It is evident that the pseudorandom set had the greatest number of outliers, i.e. six. This was followed by the 5% set with four outliers. The 15% and 25% sets had the fewest outliers with only three each. This suggests that using quantum random numbers 15% and 25% of the time results in more consistent compositions. However, considering the fact that the 25% set scores lower on average, aesthetically, than the pseudorandom set leaves only the 15% set as being preferable. The experimental results therefore suggest that using quantum random numbers 15% of the time is perhaps a 'sweet spot' since it does not also result in a lower aesthetics score or fewer compositions generated.

At worst, it is not detrimental to the composing process and at best, it is beneficial. The remaining question, then, is whether it is worth using quantum random numbers assuming its 'overhead' cost compared to the usually 'built-in' pseudorandom number generators in programming languages. To answer this, we should weigh the actual cost of using quantum random numbers (e.g. 15% of the time) in a particular computationally creative system against the benefits of doing so, i.e. not having a completely pseudorandom process that is theoretically predictable and that will probably result in repeating patterns given enough time.

Even if the quantum random numbers introduce a measure of unpredictability in a computationally creative system, the domain itself may prove to be a limiting factor. For instance, in chess, it is estimated that the upper bound of reachable, unique positions is at most about $1.7 \times 10^{46}$ [16]. The subset of positions that lead to a forced mate, for example, is therefore even smaller but still an extremely large number. The use of a degree of quantum randomness therefore will help ensure more of the search space is explorable by the computer program.

V. Conclusions

Quantum random numbers obtained from nature based on the principles of quantum mechanics are believed to be 'truly' random and better than their algorithmic pseudorandom number counterparts produced by computers. This is because 'truly' random numbers are unpredictable, even in theory. Unpredictability also happens to be a meaningful quality when it comes to creativity, as exhibited by humans or computers. This is in the sense that creativity is usually 'surprising' to us. It would tend to mean historically or 'H-creative' (never seen before) rather than psychologically or 'P-creative' (new to the person) [17].

Therefore, in a computationally creative system (notably in this case one that does not employ any kind of machine learning), the use of quantum random numbers might yield benefits as well. The experimental results indicated that there does appear to be a slight advantage to using quantum random numbers (15% of the time) during the creative process. It results in only half of the statistical outliers but suffers nothing in terms of aesthetics or the quantity of compositions produced. If the quantum randomness employed in the computationally creative system requires additional resources compared to default internal pseudorandom number generation, this needs to be weighed against the overall benefits obtained.

No two computationally creative systems are alike, even if they produce the same type of output and use many of the same approaches. The applicability of the findings of this research should therefore not be taken for granted. However, 15% quantum randomness might be a good place to start or a preferable default setting, especially when the means to experimentally verify what suits a particular system best is not available. Future work in this area, depending on available resources, may include more detailed analyses of automatic chess problem generation (e.g. involving longer mates and studies) and testing with even more different frequencies of quantum randomness, e.g. 10%, 30%, 35%, 75%.

Similar research work can also be done in other domains of computational creativity such as music and art. At some point a meta-analysis of some kind can be performed to assess with more confidence the effects quantum randomness has on computationally creative systems, in general. There may even be an overlap with the effects it has on machine learning systems. It could also turn out that in the case of chess, being a fundamentally finite and mathematical domain of investigation, introducing more and more 'true' randomness into the system eventually results in a degradation of performance whereas in other domains such as paintings and photography the threshold may be higher or undetectable.